\documentclass[11pt]{article}

\usepackage[preprint]{acl}

\usepackage{times}
\usepackage{latexsym}

\usepackage[T1]{fontenc}

\usepackage[utf8]{inputenc}

\usepackage{microtype}

\usepackage{inconsolata}

\usepackage{graphicx}

\usepackage{amsmath}
\usepackage{comment}
\usepackage{dirtytalk}
\usepackage{multirow}
\usepackage{array}
\usepackage{amsfonts}
\usepackage{threeparttablex}
\usepackage{fancyvrb}
\usepackage{fvextra}

\newcommand{\pb}[1]{\vspace{0.75ex}\noindent{\bf \em #1}\hspace*{.3em}}

\title{Efficient RAG with Intent-Aware Retrieval and Semantics-Preserving Chunking}

\author{
 \textbf{Fachrina Dewi Puspitasari\textsuperscript{1}},
 \textbf{Chaoning Zhang\textsuperscript{1,*}},
 \textbf{Jiaquan Zhang\textsuperscript{1}},\\
 \textbf{Zhicheng Wang\textsuperscript{1}},
 \textbf{Hafiz Shakeel Ahmad Awan\textsuperscript{1}},
 \textbf{Rizwan Qureshi\textsuperscript{2}},
 \textbf{Jewon Lee\textsuperscript{3}},\\
 \textbf{Tae-Ho Kim\textsuperscript{3}},
 \textbf{Yang Yang\textsuperscript{1}}
\\
\\
 \textsuperscript{1}School of Computer Science and Engineering, University of Electronic Science and Technology of China,\\
 \textsuperscript{2}Massachusetts General Hospital, Harvard University,
 \textsuperscript{3}Nota AI,\\
 \small{
   \textbf{*Correspondence:} \href{mailto:chaoningzhang1990@gmail.com}{chaoningzhang1990@gmail.com}
 }
}

\begin{document}
\maketitle
\begin{abstract}
The demand for powerful instruction following and reasoning capability of large language models (LLMs) has promoted rapid development of retrieval-augmented generation (RAG).
The RAG system assists LLM generation by retrieving chunks of query-fit supplementary knowledge from an external database.
Conventional RAG systems, however, suffer from information insufficiency due to two factors, which are intent-agnostic retrieval and information fragmentation. 
Our work proposes a RAG framework, termed InSemRAG, that addresses these challenges via an iterative retrieve-and-check mechanism with two supporting modules, an intention-aware retriever (IAR) and semantics-preserving chunking (SPC).
IAR implements a dynamic hybrid retrieval method that adaptively weights the retrieval channels based on the query intent, while SPC performs detection and reparation to the damaged evidence chunks to preserve the semantic integrity.
To alleviate the computational latency brought by our iterative mechanism, we leverage small language models (SLMs).
Extensive experiments across several benchmark datasets consistently demonstrate the competitiveness of our method against recent state-of-the-art RAG mechanisms.
Particularly, our method achieves significant gains on multi-hop and evidence-sensitive tasks, with a 2.65-point improvement in F1 on HotPotQA and a 1.5-point increase in accuracy on FEVER.
Our method also achieves competitive performance to Multi-Hop RAG with 4.32$\times$ lower latency with the utilization of SLM.

\begin{figure*}[!htb]
    \centering
    \includegraphics[width=.845\linewidth]{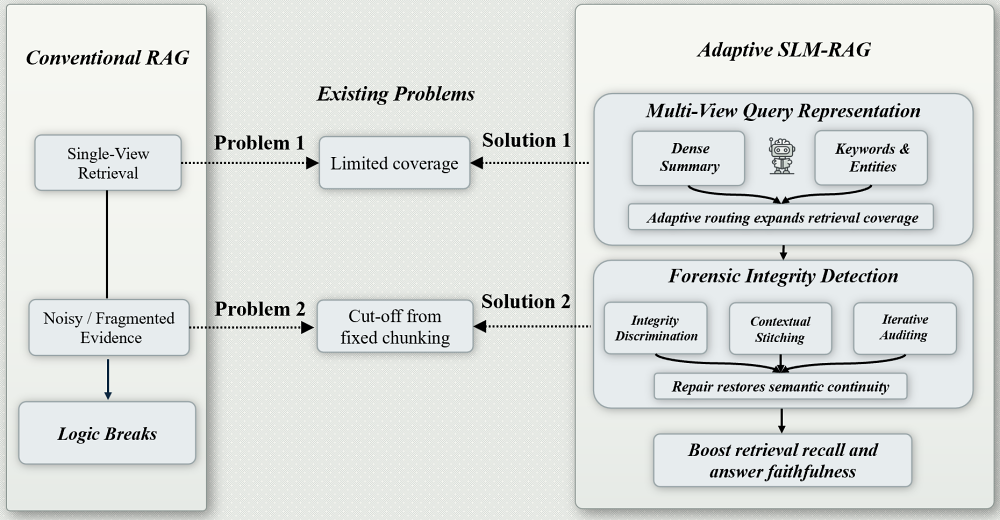}
    \caption{Limitations on the conventional RAG system that relies on a fixed retrieval channel and ignores semantically fragmented evidence chunks. We address these limitations in InSemRAG.}
    \label{fig:motivation}
\end{figure*}
\end{abstract}

\section{Introduction}


With the breakthrough of Large Language Models (LLMs) in question answering, code generation, and multimodal understanding~\citep{gao2023retrieval,wang2025think,wang2026efficient,li2026arise}, Retrieval-Augmented Generation (RAG)~\citep{lewis2020retrieval} has become the mainstream paradigm for improving model factual consistency and knowledge coverage.
RAG effectively alleviates the problems of knowledge timeliness and hallucination generation in language models by combining external knowledge base retrieval with generative models~\citep{wang2023augmenting,zhang2026tdarc}. 
Specifically, RAG systems consist of three core steps, namely retrieval, chunking, and generation.
First, the user provides an input query to the retriever, which derives retrieval keywords for sparse retrieval or rewrites the query for dense retrieval.
The retriever then extracts uniform-length chunks of information that are highly similar to the retrieval query.
The retrieved evidence is then aggregated and sent to the LLM as a generator to produce an answer to the user's query.

The standard RAG mechanisms, however, are prone to information insufficiency in the retrieved evidence.
The fixed retrieval channel (e.g., dense~\cite{borgeaud2022improving, izacard2022few, lewis2020retrieval} or sparse~\cite{ram2023context, huly2024old}) commonly leveraged in the contemporary RAG systems tends to ignore the user's requirement~\cite{wang2024evaluating}.
Such single-channel retrieval may collect descriptive evidence for semantically complex queries or retrieve explanatory evidence for purely descriptive questions, leading to evidence–intent mismatches.
Moreover, mechanical chunking (e.g., boundary-based chunking~\cite{finardi2024chronicles} or structure-aware chunking~\footnote{https://python.langchain.com}~\cite{zhang2026text}) may break the content or ignore the semantic connections, posing a risk to information fragmentation~\citep{izacard2021leveraging}.
Recent studies on efficient neural modeling and representation learning further suggest that rigid processing pipelines can hinder the preservation of temporal and structural dependencies in sequential data, motivating more adaptive and context-aware retrieval and encoding strategies~\citep{8368077,9540752}.

To address these challenges, we propose InSemRAG, a RAG framework that addresses the problem of information insufficiency in the retrieved evidence. 
%
%
%
InSemRAG consists of two modules, an intention-aware retriever (IAR) and semantics-preserving chunking (SPC).
Given an input query from the user, we first perform dual-view rewriting that simultaneously decomposes the query into keywords for sparse retrieval and expands it into a semantically meaningful phrase for dense retrieval. 
Utilizing the rewritten query, we perform dynamic hybrid retrieval, where we dynamically modulate the retrieval channel weight depending on the characteristics of the query.
Instead of passing the retrieved evidence directly to the generator, we perform detection and reparation on the semantically damaged chunk.
Specifically, we repair the damaged chunk by aggregating with the nearest neighbor chunks from the source documents and rewriting it to meet the chunk length requirement while maintaining the semantic information.
Finally, we assess whether the final evidence contains all query-required information elements.
Otherwise, the retrieve-and-check procedure is repeated until information completeness is satisfied.

To accommodate the complexity of the iterative retrieve-and-check mechanism in InSemRAG, we leverage small language models (SLMs) as the machines for retrieval and semantic checking.
We are motivated by the excellent instruction following capability at minimum computational cost brought by SLMs~\cite{wang2025comprehensive,lee2024mentor}.
This utilization circumvents the need for large instruction-following models utilized in recent RAG systems~\cite{shen2024retrieval}.

We summarize our contributions as follows:
\begin{itemize}
    \item To the best of our knowledge, we are the first to introduce a semantic integrity-based iterative dynamic retrieve-and-check mechanism in RAG.
    
    \item We propose InSemRAG, an efficient intent-aware retrieval and semantics-preserving chunking mechanism for RAG to retrieve evidence that not only considers the user's requirement, but also maintains the evidence's semantic integrity.
    
    \item Through a comprehensive experimental validation, we demonstrate the effectiveness of InSemRAG in both performance and efficiency dimensions. Specifically, we yield up to 2.65-point improvement in F1 on evidence-sensitive reasoning tasks with 4.32× lower latency than prior state-of-the-art methods.
    
\end{itemize}


\section{Related Work}

RAG~\citep{lewis2020retrieval, gao2023retrieval} addresses fundamental limitations of specific knowledge coverage in LLMs by integrating external knowledge retrieval with generative models~\citep{wang2023augmenting}.

\pb{Query Enhancement Techniques.}
Query is one of the factors that determine the success of retrieval mechanism for the generator~\cite{wang2024evaluating}.
Thus, several works explore the RAG improvement technique from the query enhancement perspective.
RQ-RAG~\citep{chan2024rq} enables query rewriting, decomposition, and disambiguation. 
DMQR-RAG~\citep{li2024dmqr} proposes diverse multi-query rewriting strategies operating at different information levels. 
HyDE~\citep{gao2022precise}  generates hypothetical documents to bridge the semantic gap between queries and relevant passages. 
Step-back Prompting~\citep{zheng2023take} rewrites detailed queries at higher conceptual levels. 
In addition, recent multimodal reasoning studies emphasize structured intermediate reasoning processes (e.g., Visual Chain-of-Thought) to enhance alignment between queries and retrieved evidence, further highlighting the importance of query abstraction and decomposition in complex reasoning tasks~\citep{zheng2026towards}.
Our method is inspired by these works, thus leveraging a query rewriting mechanism in the pre-retrieval process through both decomposition and enhancement.

\pb{Evidence Retrieval Methods.}
Contemporary RAG systems are generally grouped into two based on retrieval mechanisms, sparse and dense.
Sparse retrieval refers to utilizing keywords to look for information in the database with BM25 or TF-IDF mechanisms~\cite{ram2023context, huly2024old}.
Meanwhile, dense retrieval leverages similarity measures between the semantic representation of the user's query and the candidate evidence~\cite{borgeaud2022improving, izacard2022few, lewis2020retrieval}.
Additionally, recent works combine sparse and dense retrievals. 
RAG-Fusion~\citep{rackauckas2024rag} applies reciprocal rank fusion to combine results from multiple query reformulations.
DAT~\citep{hsu2025dat} proposes dynamic alpha tuning for hybrid retrieval, adapting weight coefficients based on individual query characteristics.
Late-interaction retrieval models such as ColBERTv2 improve retrieval granularity by preserving token-level interactions, enabling more precise matching between queries and passages~\citep{santhanam2022colbertv2}.
Beyond these paradigms, recent efforts explore more fine-grained and efficient retrieval mechanisms. For instance, HASH-RAG introduces deep hashing techniques to enable efficient yet precise retrieval~\citep{guo2025hash}, while instruction-driven overlap reduction further improves retrieval efficiency by minimizing redundant evidence~\citep{ou2025accelerating}.
Nevertheless, these approaches assume that the post-retrieval condition is perfect and thus, the evidence is directly passed to the generator.

\pb{Evidence Integrity Checking.} 
Post-retrieval mechanisms, including reranking and context compression, address evidence quality.
RankRAG~\citep{yu2024rankrag} unifies context ranking with retrieval-augmented generation, using LLMs to rank retrieved passages before feeding them to the generator. 
Late Chunking~\citep{gunther2024late} addresses semantic fragmentation by encoding documents with full in-document context before chunking, preserving contextual information that fixed-length chunking typically loses. 
Structure-aware approaches~\citep{zhang2026text} further tackle this by incorporating global document structure into the evidence processing pipeline. 
Despite the growing research in semantic chunking, the proposed methods generally incur high computational cost, which cannot be sufficiently justified by the gains proposed~\citep{qu2025semantic}.



\pb{Computationally Efficient RAG.} 
To address the increasingly complex RAG framework and the utilization of large generative models as a retriever, current studies explore ways to make LLM and RAG more efficient~\citep{zhou2026look,wang2026stream,wang2026efficient}.
MiniRAG~\citep{fan2025minirag} utilizes heterogeneous graph indexing, combining text chunks with entity hierarchies to reduce reliance on dense embeddings. 
Plan\textsuperscript{*}RAG~\citep{verma2024plan} structures multi-hop reasoning as directed acyclic graphs (DAGs), enabling systematic exploration of reasoning paths with parallel execution for efficiency. 
LightRAG~\citep{guo2024lightrag} incorporates graph structures into text indexing with dual-level retrieval, enhancing contextual awareness while improving response times.
Atlas~\citep{izacard2022atlas} demonstrates that jointly pretraining retrieval and generation at scale can significantly improve knowledge-intensive task performance while maintaining efficiency.
Recent works also investigate model compression and lightweight architectures for multimodal and language models, such as knowledge restructuring techniques that transform external information into structured triplets for more efficient retrieval~\citep{wang2026transforming} and
lightweight agent memory mechanisms built upon small language models~\citep{zhang2026lightweight}.
Furthermore, advances in efficient multimodal agent learning highlight the role of transferable experience in reducing redundant computation during reasoning~\citep{li2026experience}. 
Our work is inspired by these studies in leveraging smaller-sized yet powerful retriever to reduce the computational cost of RAG framework.

\section{Method}
This section systematically describes the overall structure and key modules of the proposed InSemRAG.
InSemRAG is designed to build an efficient and evidence-complete retrieval-augmented generation system using SLMs as a retriever.

\subsection{Preliminary}
\pb{Problem Definition.}
We formulate RAG as a sequential decision-making process under a probabilistic graph model, where given a query $q$ and a parameterized generator $P_\theta$, RAG objective is to maximize the log-likelihood $\log P(y|q)$ of the answer $y$ with the assistance of a retriever $P_\phi$ that collects relevant evidence $\mathcal{E}$ from external knowledge database $\mathcal{K}$. 

\begin{equation}
  P(y|q) \approx \sum_{\mathcal{E} \in \mathcal{E}_{top}} \underbrace{P_{\theta}(y|q, \mathcal{E})}_{\text{Generator}} \cdot \underbrace{P_{\phi}(\mathcal{E}|q, \mathcal{K})}_{\text{Retriever}}
\end{equation}

\noindent
Marginalization over all information in the external knowledge base $\mathcal{K}$ is computationally infeasible due to its size; therefore, latent evidence $\mathcal{E}$ is used to approximate $\mathcal{K}$.
Let $\mathcal{K} = \{d_1, \dots, d_m\}$ where each document $d_i$ is partitioned into $n$ set of information chunks $\{c_{1}, \dots, c_{n}\}$.
All possible combinations of these chunks are defined as the evidence space $\mathbb{E}$.
The optimal subset $\mathcal{E}^* \subset \mathbb{E}$ satisfies the following constraints:

\begin{equation}\label{eq:optimal}
\mathcal{E}^* = \underset{\mathcal{E} \in \mathbb{E}}{\arg\max} \left[ \underbrace{\mathcal{S}_{rel}(q, \mathcal{E})}_{\text{Semantic}} + \lambda \cdot \underbrace{\mathcal{C}_{int}(\mathcal{E})}_{\text{Context}} \right]
\end{equation}

\noindent
where $S_{rel}$, $C_{int}$, and $\lambda$ measure semantic alignment, contextual integrity, and the weight that determines the tradeoff between the two, respectively.

\pb{Motivation.}
Conventional RAGs entail several limitations that we illustrate in Figure~\ref{fig:motivation}:
\begin{itemize}
    \item Utilize a fixed (e.g., dense or sparse) retrieval function $R_\phi:\mathcal{Q}\xrightarrow{}S^k$ which returns top-$k$ slices $\mathcal{E}=R_\phi(q)$, where $R_\phi$ is independent of query structure.
    This may create a gap between what the generator expects and what the retriever produces, which becomes larger as the query complexity increases.
    \item Combine the evidence chunks with mechanical chunking $\Gamma:d\xrightarrow{}\{c_1, \dots, c_n\}$ that merges all chunks at a fixed-length $l$, thus truncating long chunks. 
    This risks discarding essential information that may exist in the truncated part, breaking the logical dependencies across chunks.
\end{itemize}

\noindent
Our work addresses these limitations by reformulating the RAG mechanisms as adaptive-corrective mechanisms.


\subsection{Technical Framework}
\begin{figure*}
    \centering
    \includegraphics[width=\linewidth]{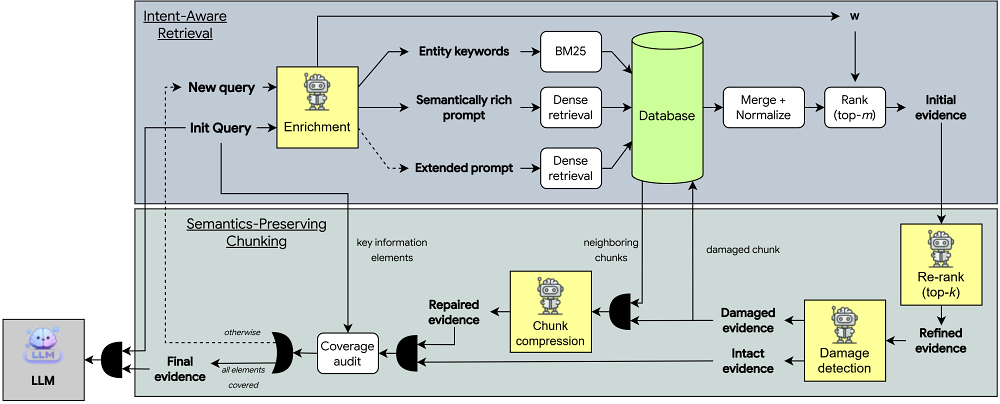}
    \caption{Illustration of InSemRAG framework. 
    In a high-level view, we first enrich the input query from multiple views, from which we retrieve the evidence through weighted multi-channel retrieval. After reranking, we perform detection on damaged evidence chunks and repair via semantic stitching. We implement our method iteratively using SLM (yellow module in the figure) for efficient computation.}
    \label{fig:framework}
\end{figure*}

We propose an intent-aware retrieval and semantics-preserving chunking in the RAG system that utilizes SLM as retriever.
Our proposed system consists of two core modules: intent-aware retrieval (IAR) and semantics-preserving chunking (SPC).
Figure~\ref{fig:framework} illustrates our framework.

\subsubsection{Query Intent-Adaptive Retrieval}
We propose IAR to address the problem of fixed retrieval that tends to ignore the characteristics of query.
We perform this retrieval through three mechanisms: query enriching, dynamic channel weighting, and hybrid-weighted retrieval.

\pb{Query Enrichment.}
User queries are often imperfect.
They can be too short, thus lacking information, or too long, thus containing irrelevant content.
These characteristics create a distribution shift that jeopardizes the distance between true requirement implied in the question and the evidence candidates.
To alleviate the issue, we propose to project query $q$ into a richer form $p$ that has the highest distribution similarity with $\mathcal{E}$ with SLM as the projector $\mathcal{P}$.
\begin{equation}
p = \mathcal{P}(q), \hspace{10pt} \mathcal{P}:\mathcal{Q} \xrightarrow{}\mathcal{P}
\end{equation}
where $p$ consists of two orthogonal vectors $p_d\perp p_s$.
This decomposition aims to balance between the semantic coverage and lexical fidelity.
The $\{p^{d}_i\}_{i=1}^m \in P_d$ defines a \textit{dense intermediary subspace} that captures semantic intent.
SLM generates the subspace by performing semantic completion through query paraphrasing to project $q$ to the clustering center.
On the other hand, $\{p^{s}_j\}_{j=1}^n \in P_s$ defines a \textit{sparse intermediary subspace} that captures entity constraints.
To generate $P_s$, SLM extracts keywords, named entities, and rare terms to match long-tail low-frequency words.

\begin{equation}
P = SLM(Q), \hspace{10pt} \text{where} \hspace{5pt} \{P_s,P_d\} \in P
\end{equation}

\pb{Dynamic Channel Weighting.}
Different queries $q$ may have different characteristics that influence how the retrieval should be performed.
Some queries are more specific, such as \say{Where does SpongeBob live?}, or more abstract \say{How to avoid rejection on conference submission?}, or require reasoning \say{Why the work-from-home trend started to boom in 2020?}, or a combination of them.
We argue that retrieval mechanisms shall adapt to these variations. 
To this end, we propose to condition retrieval channels on the query $q$, by performing soft weighting that dynamically control the contribution of each retrieval channel.
Specifically, we derive the weight $\textbf{w}$ from SLM through the following operation.

\begin{equation}
\textbf{w}=\text{Softmax}\left(\text{MLP}\left(\text{SLM}\left(q\right)\right)\right)
\end{equation}

\noindent
where $\textbf{w}=\left[\alpha,\beta,\gamma\right]^T$ and $\alpha +\beta+\gamma=1$. 
These weights control the contributions of dense, sparse, and extended retrievals, respectively.
We provide more analysis on these weight factors in Appendix~\ref{sec:weight}.

\pb{Hybrid Retrieval with Weighted Fusion.}
Using two forms of enrich query $p_d$ and $p_s$, we perform both dense and sparse retrieval in parallel.
In dense retrieval, we embed both dense query $p_d$ and the document $d$ using the same encoder $\phi(\cdot)$ to compute the cosine similarity between the two.
Thus, the score of candidate evidence from dense $S_d$ and extended retrieval is obtained through the following formula.

\begin{equation}
S_d=\underset{p_d \in P_d}{\max} \hspace{5pt} \cos \left(\phi\left(p_d\right), \phi\left(d\right)\right)
\end{equation}

\noindent
Meanwhile, the sparse retrieval process utilizes the BM25~\cite{robertson2009probabilistic} algorithm and the score $S_s$ aggregates the contribution of all keywords as follow.

\begin{equation}
S_s=\underset{p_s \in P_s}{\max} \hspace{5pt} \text{BM25} \left(p_s, d\right)
\end{equation}

\noindent
We merge and normalize all scores $\{\hat{S}_d, \hat{S}_s, \hat{S}_e\} \in \hat{S}$, then apply the weight $\textbf{w}$ to get final score for ranking $S$. 

\begin{equation}
S=\textbf{w}^\text{T} \cdot \left[\hat{S_d}, \hat{S_s}, \hat{S}_e\right]
\end{equation}

\noindent
We then retrieve top-$m$ slices as the initial evidence set $\mathcal{E}_0=\{c_1, \dots, c_m\}$.

\subsubsection{Semantic-Preserving Chunking}\label{sec:SPC}
To address the problem of evidence truncation, we propose a chunk stitching mechanism that preserves the semantic of the chunks, termed SPC.

\pb{Candidate Refinement.}
The defined $m$ of the initial evidence set $\mathcal{E}_0$ is large to allow more evidence to be retrieved.
We further refine this retrieval using SLM and the original query $q$ as the input.
Specifically, the SLM analyzes each chunk and outputs a new score based on the original query.
We utilize the new score to rerank the chunks and retrieve top-$k$ where $k < m$ as the refined evidence set $\mathcal{E}_r=\{c_1, \dots, c_k\}$.

\pb{Damage Detection.}
Given the refined evidence set $\mathcal{E}_r$, we indicate the semantic completeness of each slice using SLM as follow.

\begin{equation}
    \mathcal{I}(c) = \mathbb{I}\left[\text{SLM}(c) < \delta\right]
\end{equation}

\noindent
where $\mathbb{I}(\cdot)$ is a binary function that classifies the chunk $c$ with respect to the completeness threshold $\delta$.
The SLM determines the chunk completeness based on several factors such as syntactic cut-off, unresolved co-reference, and broken logic.
This process partitions the refined query into intact and damaged subsets $\mathcal{E}_r=\mathcal{E}_{int} \cup\mathcal{E}_{dam}$. 

\pb{Chunk Repair.}
For each chunk $c$ in $\mathcal{E}_{dam}$, we repair its semantic integrity by backtracking to the original document $d$ from which the chunk is retrieved.
We then retrieve chunks preceding $c_{pre}$ and succeeding $c_{post}$ the chunk $c$, and stitch them together.
Instead of plain expansion, we perform stitching by compression using SLM.
This is to avoid unintentional truncation due to limited LLM context window.
We maximize the information density of the compressed chunk to preserve the original semantic content from the combined chunks.
Specifically, we define this operation as follow.

\begin{equation}
    c^* = \text{SLM}_{compress}(c_{pre} \oplus c \oplus c_{post}, q)
\end{equation} 

\noindent
where $c^*$ defines the repaired chunk.

\pb{Coverage Auditing.}
The chunk repair process produces a repaired evidence set $\mathcal{E}_{rep}$.
We further combine this repaired set with the intact set as a new evidence set $\mathcal{E}_t =\mathcal{E}_{int} \cup\mathcal{E}_{rep}$.
We use this set and perform auditing to check whether this new evidence set is semantically sufficient to answer the query.
Specifically, we identify the discrepancy in the semantic completeness $\Delta_t$ by first extracting key information elements $e$ from the query $\text{KIE}(q)$, and measure the entailment score $\text{Ent}(\cdot)$ between each element $e$ and the evidence set $\mathcal{E}_t$.

\begin{equation}\label{eq:kie}
    \Delta_t = \{e \in \text{KIE}(q) \mid \text{Ent}(\mathcal{E}_t, e) < \epsilon\}
\end{equation}

\noindent
If $\Delta_t \neq \emptyset$, we convert the corresponding $e$ into a new query $q_{t+1}$ that will trigger a new round of RAG, repeating the process of IAR and SPC iteratively.
We provide analysis on this iterative behavior in Appendix~\ref{sec:iter}.

\begin{table*}[!htb]
\small
    \centering
    \begin{threeparttable}
    \begin{tabular}{|p{45pt}|c|c|c|c|c|c|c|c|p{30pt}|p{30pt}|}
    \hline
        \multirow{2}*{\raisebox{-0.5\height}{\textbf{Method}}} & \multirow{2}*{\raisebox{-0.5\height}{\textbf{LLM}}} & \textbf{NQ-open} & \multicolumn{2}{c|} {\textbf{TriviaQA}} & \multicolumn{2}{c|}{\textbf{WebQuestions}} & \multicolumn{2}{c|}{\textbf{HotpotQA}} & \multicolumn{2}{c|}{\textbf{2WikiMultiHopQA}} \\
        \cline{3-11}
        && EM & EM & F1 & EM & F1 & EM & F1 & EM & F1 \\
        \hline
        \multirow{3}{40pt}{w/o RAG} & GPT & 35.06 & 44.38 & 48.15 & 37.43 & 45.81 & 28.41 & 34.68 & 41.63 & 49.13 \\
        & Qwen & 33.31 & 42.16 & 45.74 & 35.56 & 43.52 & 25.57 & 31.21 & 37.47 & 44.22 \\
        & DeeepSeek & 36.11 & 45.71 & 49.59 & 38.55 & 47.18 & 29.83 & 36.41 & 43.71 & 51.59 \\
        \hline
        \multirow{3}{40pt}{Na\"ive RAG} & GPT & 48.35 & 58.31 & 63.52 & 44.52 & 57.63 & 39.14 & 44.85 & 52.33 & 57.71 \\
        & Qwen & 39.78 & 55.39 & 60.34 & 42.29 & 54.75 & 35.23 & 40.36 & 47.1 & 51.94 \\
        & DeeepSeek & 43.13 & 60.06 & 65.43 & 45.86 & 59.36 & 41.1 & 47.09 & 54.95 & 60.6 \\
        \hline
        \multirow{3}{40pt}{CoT-on-Concat} & GPT & 49.25 & 60.57 & 67.38 & 44.91 & 56.42 & 39.36 & 45.38 & 52.55 & 58.24\\
        & Qwen & 41.96 & 57.54 & 64.01 & 42.66 & 53.6 & 35.42 & 40.84 & 47.3 & 52.42\\
        & DeeepSeek & 45.5 & 62.39 & 69.4 & 46.26 & 58.11 & 41.33 & 47.65 & 55.18 & 61.15\\
        \hline
        \multirow{3}{40pt}{Wikipedia Graph} & GPT & 49.68 & 54.81 & 59.71 & 42.37 & 55.88 & 51.76 & 60.14 & 55.52 & 64.51\\
        & Qwen & 42.17 & 52.07 & 56.72 & 40.25 & 53.09 & 46.58 & 54.13 & 49.97 & 58.06\\
        & DeeepSeek & 45.72 & 56.45 & 61.5 & 43.64 & 57.56 & 54.35 & 63.15 & 58.3 & 67.74\\
        \hline
        \multirow{3}{40pt}{FiD} & GPT & 53.62 & 72.61 & 76.51 & 49.30 & 62.81 & 56.10 & 62.51 & 56.14 & 65.62\\
        & Qwen & 50.94 & 68.98 & 72.68 & 46.83 & 59.67 & 50.49 & 56.26 & 50.53 & 59.06\\
        & DeeepSeek & 55.23 & 74.79 & 78.81 & 50.78 & 64.69 & 58.9 & 65.64 & 58.95 & 68.9\\
        \hline
        \multirow{3}{40pt}{ReAct} & GPT & 51.45 & 63.97 & 74.24 & 42.91 & 55.14 & 56.27 & 64.20 & 56.52 & 66.52\\
        & Qwen & 47.51 & 60.77 & 70.53 & 40.76 & 52.38 & 50.64 & 57.78 & 50.87 & 59.87\\
        & DeeepSeek & 51.51 & 65.89 & 76.47 & 44.2 & 56.79 & 59.08 & 67.41 & 59.35 & 69.85\\
        \hline
        \multirow{3}{40pt}{IRCoT} & GPT & 52.3 & 69.21 & 75.04 & 46.36 & 59.62 & 44.51 & 55.14 & 57.07 & 68.32\\
        & Qwen & 49.68 & 65.75 & 71.29 & 44.04 & 56.64 & 40.06 & 49.63 & 51.36 & 61.49\\
        & DeeepSeek & 53.87 & 71.29 & 77.29 & 47.75 & 61.41 & 46.74 & 57.9 & 59.92 & 71.74\\
        \hline
        \multirow{3}{40pt}{SELF-RAG} & GPT & 53.14 & 70.38 & 78.88 & 48.96 & 61.58 & 43.61 & 51.79 & 58.71 & 69.50\\
        & Qwen & 50.48 & 66.86 & 74.94 & 46.51 & 58.5 & 39.25 & 46.61 & 52.84 & 62.55\\
        & DeeepSeek & 54.73 & 72.49 & 81.25 & 50.43 & 63.43 & 45.79 & 54.38 & 61.65 & 72.98\\
        \hline
        \multirow{3}{40pt}{Multi-Hop RAG} & GPT & 50.15 & 68.76 & 74.81 & 43.12 & 56.97 & 54.17 & 63.42 & 56.91 & 66.35\\
        & Qwen & 45.89 & 65.32 & 71.07 & 40.96 & 54.12 & 48.75 & 57.08 & 51.22 & 59.72\\
        & DeeepSeek & 49.76 & 70.82 & 77.05 & 44.41 & 58.68 & 56.88 & 66.59 & 59.76 & 69.67\\
        \hline
        \multirow{3}{40pt}{\textbf{InSemRAG (Ours)}} & GPT & \textbf{53.85} & \textbf{71.12} & \textbf{79.45} & \textbf{50.15} & \textbf{63.24} & \textbf{57.42} & \textbf{66.85} & \textbf{59.85} & \textbf{70.65}\\
        & Qwen & \textbf{51.25} & \textbf{67.95} & \textbf{75.82} & \textbf{47.65} & \textbf{59.85} & \textbf{52.15} & \textbf{59.45} & \textbf{53.45} & \textbf{63.85}\\
        & DeeepSeek & \textbf{55.65} & \textbf{73.15} & \textbf{81.85} & \textbf{51.55} & \textbf{65.15} & \textbf{60.45} & \textbf{68.95} & \textbf{62.85} & \textbf{74.15}\\
        \hline
    \end{tabular}
    \caption{RAG evaluation on question-answering benchmark dataset.}
    \label{tab:main1}
    \end{threeparttable}
\end{table*}

\section{Results}

This section elaborates on the evaluation of our SLM-RAG method.
We utilize Llama-3.2-1B-Instruct as SLM.
This small-sized language model is sufficient to function as a semantic controller for all related subtasks in our method.
This benefit arises more from the instruction-following consistency than the model scale.
Unless otherwise specified, we set retrieval depths at $m=20$ and $k=10$.
Additionally, we set completeness threshold at $\delta=0.5$ and coverage threshold at $\epsilon=0.6$ (further details in Appendix~\ref{sec:threshold}).

\begin{table*}[!htb]
\small
    \centering
    \begin{threeparttable}
    \begin{tabular}{|p{50pt}|c|c|c|c|c|c|c|}
    \hline
        \multirow{2}*{\raisebox{-0.5\height}{\textbf{Method}}} & \multirow{2}*{\raisebox{-0.5\height}{\textbf{LLM}}} & \multicolumn{3}{c|} {\textbf{ELI5}} & \multicolumn{3}{c|}{\textbf{FEVER}} \\
        \cline{3-8}
        && ROUGE-1 & ROUGE-2 & ROUGE-L & Acc & FEVER Score & F1 \\
        \hline
        \multirow{3}{50pt}{w/o RAG} & GPT & 23.62 & 4.25 & 20.51 & 61.97 & 50.87 & 64.52\\
        & Qwen & 21.73 & 3.91 & 18.87 & 58.87 & 48.33 & 61.29 \\
        & DeepSeek & 23.15 & 4.16 & 20.1 & 63.83 & 52.4 & 66.46 \\
        \hline
        \multirow{3}{40pt}{Na\"ive RAG} & GPT & 30.21 & 6.77 & 24.32 & 66.82 & 62.39 & 77.17 \\
        & Qwen & 27.79 & 6.23 & 22.37 & 63.48 & 59.27 & 73.31 \\
        & DeepSeek & 29.61 & 6.63 & 23.83 & 68.82 & 64.26 & 79.49 \\
        \hline
        \multirow{3}{40pt}{CoT-on-Concat} & GPT & 31.20 & 8.39 & 25.31 & 68.91 & 66.34 & 78.35 \\
        & Qwen & 28.7 & 7.72 & 23.29 & 65.46 & 63.02 & 74.43 \\
        & DeepSeek & 30.58 & 8.22 & 24.8 & 70.98 & 68.33 & 80.7 \\
        \hline
        \multirow{3}{40pt}{Wikipedia Graph} & GPT & 32.01 & 9.15 & 26.91 & 74.39 & 67.31 & 80.76\\
        & Qwen & 29.45 & 8.42 & 24.76 & 70.67 & 63.94 & 76.72\\
        & DeepSeek & 31.37 & 8.97 & 26.37 & 76.62 & 69.33 & 83.18\\
        \hline
        \multirow{3}{40pt}{FiD} & GPT & 34.72 & 10.52 & 27.37 & 80.09 & 71.39 & 86.31\\
        & Qwen & 31.94 & 9.68 & 25.18 & 76.09 & 67.82 & 81.99\\
        & DeepSeek & 34.03 & 10.31 & 26.82 & 82.49 & 73.53 & 88.9\\
        \hline
        \multirow{3}{40pt}{ReAct} & GPT & 30.21 & 7.26 & 24.81 & 77.52 & 69.37 & 83.03\\
        & Qwen & 27.79 & 6.68 & 22.83 & 73.64 & 65.9 & 78.88\\
        & DeepSeek & 29.61 & 7.11 & 24.31 & 79.85 & 71.45 & 85.52\\
        \hline
        \multirow{3}{40pt}{IRCoT} & GPT & 31.28 & 9.63 & 27.34 & 82.61 & 74.28 & 87.01\\
        & Qwen & 28.78 & 8.86 & 25.15 & 78.48 & 70.57 & 82.66\\
        & DeepSeek & 30.65 & 9.44 & 26.79 & 85.09 & 76.51 & 89.62\\
        \hline
        \multirow{3}{40pt}{SELF-RAG} & GPT & 34.09 & 12.83 & 30.20 & 83.62 & 79.68 & 89.92\\
        & Qwen & 31.36 & 11.8 & 27.78 & 79.44 & 75.7 & 85.42\\
        & DeepSeek & 33.41 & 12.57 & 29.6 & 86.13 & 82.07 & 92.62\\
        \hline
        \multirow{3}{40pt}{Multi-Hop RAG} & GPT & 27.86 & 6.32 & 23.61 & 73.62 & 68.67 & 82.00\\
        & Qwen & 25.63 & 5.81 & 21.72 & 69.94 & 65.24 & 77.9\\
        & DeepSeek & 27.3 & 6.19 & 23.14 & 75.83 & 70.73 & 84.46\\
        \hline
        \multirow{3}{40pt}{\textbf{InSemRAG (Ours)}} & GPT & \textbf{35.42} & \textbf{13.45} & \textbf{31.15} & \textbf{85.12} & \textbf{81.25} & \textbf{91.05}\\
        & Qwen & \textbf{32.85} & \textbf{12.45} & \textbf{28.95} & \textbf{81.15} & \textbf{77.25} & \textbf{86.85}\\
        & DeepSeek & \textbf{34.85} & \textbf{13.15} & \textbf{30.85} & \textbf{87.45} & \textbf{83.65} & \textbf{93.55}\\
        \hline
    \end{tabular}
    \caption{RAG evaluation on long-form generation and fact verification benchmarks. Our method yields competitive performance over strong reasoning-centric baselines.}
    \label{tab:main2}
    \end{threeparttable}
\end{table*}

\subsection{Main Results}
We evaluate the RAG performance on three LLMs as generators.
Specifically, we utilize GPT-4o-mini, Qwen-turbo, and DeepSeek-V3 to cover diverse architectures and deployment settings, enabling a fair evaluation of robustness and cross-model generalization.
We compare our method against several strong baselines such as generation without RAG, Na\"ive RAG~\citep{lewis2020retrieval, gao2023retrieval}, CoT-on-Concat~\cite{wei2022chain}, Wikipedia Graph~\cite{lin2025structured}, FiD~\cite{izacard2021leveraging}, ReAct~\cite{yao2022react}, IRCoT~\cite{trivedi2023interleaving}, SELF-RAG~\cite{asai2024self}, and Multi-Hop RAG~\cite{tang2024multihop}.
We evaluate on five question-answering (QA) benchmark datasets (i.e, NQ-open~\cite{kwiatkowski2019natural}, TriviaQA~\cite{joshi2017triviaqa}, WebQuestions~\cite{berant2013semantic}, HotpotQA~\cite{yang2018hotpotqa}, and 2WikiMultiHopQA~\cite{ho2020constructing}), long-form generation benchmark (ELI5~\cite{fan2019eli5}), and fact verification benchmark (FEVER~\cite{thorne2018fever}).
We measure Exact Match (EM) and F1 scores for QA tasks, ROUGE scores~\cite{lin2004rouge} for long-form generation task, as well as accuracy, FEVER Score~\cite{thorne2018fever}, and F1 score on fact verification task.
Table~\ref{tab:main1} presents the results on QA tasks and Table~\ref{tab:main2} shows the evaluation on long-form generation and fact verification tasks.
We observe that across all generators, our method consistently outperforms Na\"ive RAG and CoT-on-Concat on both single- and multi-hop QA, demonstrating that dual-view query enrichment and dynamic sparse–dense routing improve evidence recall without retriever retraining. 
These benefits are most pronounced on entity-centric benchmarks (NQ-open, TriviaQA, WebQuestions), where lexical precision and semantic intent must be jointly balanced. 
Performance gains further amplify on multi-hop and evidence-sensitive tasks (HotpotQA, 2WikiMultiHopQA, ELI5, FEVER), where fixed-size chunking often causes semantic fragmentation. 
In these settings, integrity-aware evidence construction yields substantial improvements over strong reasoning-centric baselines (ReAct, IRCoT, SELF-RAG). 
Consistent gains on FEVER and long-form generation also indicate improved evidential faithfulness and global coherence. 

\subsection{Ablation Studies}

\pb{Generalizability and Modular Contribution.}
To evaluate the generalizability of our framework, we swap our default SLM with Qwen2.5-1.5B-Instruct (SLM swap).
We also examine the contribution of each module to the RAG performance.
Specifically, we evaluate four modular conditioning: without query enrichment in IAR (w/o QE-IAR), without dynamic weighting (weighting retrieval channels with uniform weights $\alpha=\beta=\gamma$) in IAR (w/o DW-IAR), without candidate evidence refinement step in SPC (w/o ER-SPC), without SPC (passing the retrieved output of QAR module to LLM) (w/o SPC), and replacing SPC with heuristic rules (SPC $\xrightarrow{}$ heuristic).
Note that we utilize the following heuristic rules to determine the completeness of a chunk:
\begin{itemize}
    \item is terminated by terminal punctuation marks (i.e., ., ?, or !),
    \item contains balanced pairs of quotation marks, parentheses, and brackets, and
    \item has a minimum token length of 50.
\end{itemize}
We also test the effectiveness of our first retrieval depth value $m$ with respect to the final depth $k$ (depth $m=4k$).
Table~\ref{tab:genmod} demonstrates that there exists a significant performance gap between baseline and heuristic-based RAG (e.g., -5.60 on HotpotQA), demonstrating that semantic fragmentation often occurs behind grammatically "correct" endings, such as unresolved co-references (e.g., "He then decided...") or dangling logical premises, where SLM offers proper assistance in repairing.
Removing dynamic channel weighting in IAR consistently degrades performance, validating its role in robust evidence recall. 
Disabling query enhancement in IAR further reduces accuracy, particularly for multi-hop and abstract queries. 
The largest drops occur when SPC is removed, highlighting semantic fragmentation as a primary bottleneck in complex reasoning and long-form generation.
Bypassing candidate evidence refinement in SPC or replacing SPC with heuristic rules also significantly harms performance, confirming the necessity of semantic filtering and repair. 
Increasing $m$ yields diminishing returns, indicating that selection effectively balances recall and noise. 
Finally, similar results with alternative lightweight SLMs verify that the framework is model-agnostic. 
Overall, IAR governs what evidence is retrieved, while SPC governs how it is filtered and structured, and all components are jointly necessary for full performance gains.

\begin{table}[!htb]
\small
    \centering
    \begin{tabular}{|c|c|c|}
    \hline
        \multirow{2}*{\textbf{Evaluation}} & \textbf{HotpotQA} & \textbf{ELI5} \\
        & \textbf{(FI)} & \textbf{(ROUGE-L)} \\
        \hline
        Baseline (Ours) & 66.85 & 31.15 \\
        \hline
        SLM swap & 65.42 & 30.22 \\
        w/o QE-IAR & 64.95 & 30.25 \\
        w/o DW-IAR & 64.1 & 29.8 \\
        w/o ER-SPC & 63.80 & 28.95 \\
        w/o SPC & 59.45 & 25.12 \\
        SPS $\xrightarrow{}$ heuristic & 61.25 & 27.8\\
        retrieval depth $m=4k$ & 66.20 & 31.05\\
        \hline
    \end{tabular}
    \caption{Evaluation on generalizability and modular contribution. Both IAR and SPC make substantial contributions to performance improvement.}
    \label{tab:genmod}
\end{table}

\pb{Robustness of Hyperparameter.}
We examine the model performance under varying chunk length $l$ and final retrieval depth $k$.
We evaluate on HotpotQA and measure the F1 score.
Table~\ref{tab:hyper} demonstrates that our method is persistently stronger than other RAG frameworks across chunk length and final retrieval depth.
This result demonstrates that semantic fragmentation severely degrades conventional RAG performance as chunk sizes shrink, while our method remains robust.
Na\"ive RAG and Multi-Hop RAG deteriorate sharply under small chunks, indicating that iterative reasoning cannot offset truncated evidence. 
In contrast, our approach consistently achieves the best F1 across all settings, with the performance gap widening as fragmentation increases. 
This demonstrates that integrity-aware evidence construction effectively repairs fragmented evidence and enables robust reasoning under adversarial retrieval conditions.

\begin{table}[!htb]
\small
    \centering
    \begin{tabular}{|m{40pt}|c|c|c|c|}
    \hline
        \multirow{2}*{\textbf{Method}} & \multirow{2}*{\textbf{$l$}} & \multicolumn{3}{c|}{\textbf{$k$}} \\
        \cline{3-5}
        && \textbf{5} & \textbf{10} & \textbf{15} \\
        \hline
        \multirow{4}{0pt}{Baseline (Ours)} & \textbf{512} & 58.52 & 66.85 & 68.45 \\
        & \textbf{256} & 56.8 & 65.15 & 67.05 \\
        & \textbf{128} & 51.45 & 60.85 & 63.4\\
        & \textbf{64} & 45.25 & 54.95 & 58.1\\
        \hline
        \multirow{4}{40pt}{Na\"ive} & \textbf{512} & 48.35 & 57.71 & 59.12 \\
        & \textbf{256} & 40.12 & 48.55 & 50.4 \\
        & \textbf{128} & 32.45 & 40.15 & 43.2\\
        & \textbf{64} & 24.15 & 31.8 & 34.55\\
        \hline
        \multirow{4}{40pt}{Multi-Hop RAG} & \textbf{512} & 54.17 & 63.42 & 65.2 \\
        & \textbf{256} & 50.25 & 60.12 & 62.45 \\
        & \textbf{128} & 45.6 & 55.45 & 58.12\\
        & \textbf{64} & 38.45 & 48.2 & 51.3\\
        \hline
    \end{tabular}
    \caption{Evaluation on robustness of hyperparameters. Our method is consistently competitive against other methods across various chunk length $l$ and final retrieval depth $k$.}
    \label{tab:hyper}
\end{table}

\pb{Latency Analysis.}
To evaluate the applicability of our framework to a real-case RAG process, we measure the latency of the end-to-end process on the HotpotQA benchmark.
For fair comparison, we set a fixed chunk length of 256.
Table~\ref{tab:latent} shows that our method operates at a competitive computation latency compared to other strong baselines.
Our method only slightly increases per-query latency over Na\"ive RAG while substantially outperforming it in F1, and remains far more efficient than Multi-Hop RAG, which is both slower and less accurate. 
This demonstrates that lightweight SLM-based coordination enables robust retrieval and evidence repair without costly multi-step reasoning.

\begin{table}[!htb]
\small
    \centering
    \begin{tabular}{|c|c|c|}
    \hline
        \textbf{Method} & \textbf{F1 Score} & \textbf{Latency (s/q)} \\
        \hline
        Baseline (Ours) & 66.85 & 1.95 \\
        \hline
        Na\"ive RAG & 44.85 & 1.25 \\
        Multi-Hop RAG & 63.42 & 8.42 \\
        \hline
    \end{tabular}
    \caption{Evaluation on end-to-end latency. Our method offers competitive performance at 4$\times$ lower latency.}
    \label{tab:latent}
\end{table}

\section{Conclusion}
We propose intent-aware retrieval and semantics-prethat preserves the semantic integrity of the retrieved evidence to alleviate the problem of incomplete information retrieval in conventional RAG systems.
We utilize SLM as our core retriever for fast and proper instruction following capability.
Our work yields competitive results against state-of-the-art RAG systems at a much lower latency.
These results broaden the exploration of efficient semantic-aware RAG for LLM.

\section{Limitation}
Our InSemRAG involves a locality assumption where we conjecture that the complete referent knowledge is located within the neighborhood of the retrieved chunk.
This influences our SPC design, where we repair the semantically fragmented chunk by backtracking within its neighborhood.
In some cases, referent knowledge may be located beyond the neighborhood of the retrieved chunk.
Thus, simple backtracking to neighboring chunks may fail to recover missing semantics.
Note that the SPC module only intentionally searches the neighboring information during backtracking under the following assumptions:
\begin{itemize}
    \item locality assumption where evidence is only utilized to support local reasoning,
    \item expansion of the backtracking window may unintentionally lead to the retrieval of irrelevant information, and
    \item SPC is utilized as an evidence repair mechanism, instead of document-level co-reference resolution.
\end{itemize}
When backtracking fails to solve the semantic fragmentation, InSemRAG relies on the second level of auditing (coverage audit) to trigger new retrieval round.

\bibliography{custom}

\clearpage

\appendix

\section{Appendix}\label{sec:appendix}
\setcounter{table}{0}
\renewcommand{\thetable}{A\arabic{table}}
\subsection{SPC Repair Analysis}

\subsubsection{Definition of Semantic Fragmentation}
We define \say{semantic fragmentation} as the following situation. 
\textit{The retrieved text block that is syntactically complete, but semantically insufficient to independently support the information required by the user's query.} 
Semantic fragmentation is not equivalent to text truncation or grammatical incompleteness.
In actual RAG scenarios, a large number of evidence blocks satisfy common syntactic integrity heuristics (e.g., ending with a terminating punctuation mark, sufficient length), but still lack key entities, preconditions, or logical relationships, thus preventing the generative model from completing reliable reasoning based on this evidence (see Section~\ref{sec:SPC}).

\subsubsection{Heuristic Integrity Rule and Its Limitations}
In conventional RAG systems, the semantic integrity of chunks is typically determined by the following  heuristic rules:
\begin{itemize}
    \item Text ends with a terminal punctuation mark (e.g., \say{.}, \say{?}, \say{!});
    \item Text contains paired quotation marks, parentheses, or other symbols;
    \item Text length exceeds the minimum token threshold.
\end{itemize}
These rules effectively filter clearly truncated text blocks, but they only focus on surface syntactic structures and cannot identify hidden semantic gaps. 
From the ablation experiment (Table~\ref{tab:genmod}), we observe that even when evidence blocks meet the above heuristic conditions, semantic fragmentation still occurs frequently and significantly affects the performance of multi-hop reasoning and evidence-sensitive tasks.

\subsubsection{Example of Semantic Fragmentation and SPC Repair}
We conduct manual analysis on HotpotQA and ELI5 on several types of semantic fragmentation with their corresponding SPC repair examples.

\pb{Unresolved Co-Reference.}
Original chunk:
\textit{He then announced his resignation shortly after the investigation concluded}. 
The heuristic rule will judge this text as semantically complete for being syntactically correct and having sufficient length.
Meanwhile, SPC judges the text as having a confidence level of semantic integrity output by SLM as lower than the threshold $\delta=0.5$.
SPC will judge the text as semantically impaired because there is an unresolved pronoun \say{He} and the referent cannot be determined within the current chunk. 
The text repaired by SPC is as follows: \textit{After weeks of public pressure and an investigation into financial misconduct, Prime Minister John Smith announced his resignation}.
The repaired text restored the reference object and thus can independently support reasoning of \say{who and why}.

\pb{Truncated Premise.}
Original chunk: 
\textit{As a result, the policy was widely criticized by experts.}
SPC will give a confidence level of $\delta=0.5$ due to the absence of the key premise needed for the conclusion.
The text repaired by SPC is as follows:
\textit{After the government abruptly raised fuel taxes, the policy was widely criticized by experts for increasing living costs}.

\pb{Broken Causal.}
Original chunk: 
\textit{This led to a significant shift in public opinion.}
SPC will judge this text as fragmented because the indicator \say{This} lacks reference, and thus, creating inconsistent semantic relationship.
The text repaired by SPC is as follows:
\textit{The prolonged economic downturn led to a significant shift in public opinion against the ruling party.}

\pb{Interrupted Argument.}
Original chunk: 
\textit{The main reasons include economic pressure, social factors.}
SPC will judge this text as fragmented because the enumeration is not expanded, and the information granularity is insufficient to support reasoning.
The text repaired by SPC is as follows:
\textit{The main reasons include rising economic pressure from inflation and growing social dissatisfaction with labor policies.}

\subsubsection{Analysis on Multi-Level Thresholding}\label{sec:threshold}
The SPC module leverages two levels of semantic auditing, chunk damage detection, and evidence coverage audit.
Chunk-level detection (with $\delta$ threshold) is used to determine whether a single evidence chunk is semantically complete.
In practice, SLM outputs a semantic completeness confidence level score in the range of ([0,1]) for each chunk.
When this $\delta < 0.5$ the chunk is determined to be semantically impaired, triggering the local backtracking repair process of SPC.
Meanwhile, evidence coverage audit (with $\epsilon$ threshold) is utilized to determine whether the current evidence set covers the key information elements required by the query (see Equation~\ref{eq:kie}). 
If $\epsilon < 0.6$, then the element is considered not covered and in the current evidence set and is converted into a new query to trigger the next round of RAG.
These mechanisms ensure reparation of local semantic loss through damage detection at the chunk level and supplementation of global information insufficiency through a new retrieval round.

\subsection{Iteration and Stopping Conditions}\label{sec:iter}
InSemRAG leverages an iterative retrieve-and-check mechanism, performing semantic repair to ensure complete information supplementation.
In each iteration, InSemRAG performs the following operations:
\begin{itemize}
    \item retrieves evidence chunk based on input query, then screens the candidate evidence,
    \item leverages SPC to repair damaged evidence chunks,
    \item determines whether the evidence covers key information elements required to support the query,
    \item if some elements are not covered in the evidence, InSemRAG will convert those into new queries to trigger next iteration.
\end{itemize}

\pb{Empirical Analysis.}
We evaluate the actual iterative behavior of InSemRAG on HotpotQA and ELI5 benchmarks.
Table~\ref{tab:iter} presents the distribution of samples going through the iteration series.

\begin{table}[!htb]
\small
    \centering
    \begin{tabular}{|c|c|c|c|}
    \hline
        \textbf{Dataset} & \textbf{1\textsuperscript{st} iter} & \textbf{2\textsuperscript{nd} iter} & \textbf{3\textsuperscript{rd} iter} \\
        \hline
        HotpotQA & 72\% & 23\% & 5\% \\
        ELI5 & 65\% & 27\% & 8\% \\
        \hline
    \end{tabular}
    \caption{Distribution of samples across retrieval rounds.}
    \label{tab:iter}
\end{table}

\noindent
We observe that, in general, the iterative retrieval mechanism in InSemRAG stops after one to two iterations.
Samples that need more iterations typically correspond to complex problems with highly dispersed information.
Thereby, we set the maximum iteration round to be three to control the mechanism latency.

\subsection{Analysis On Dynamic Channel Weighting}\label{sec:weight}
The IAR module passes the input query through SLM that predicts the weight of different retrieval channels $\textbf{w}=\left[\alpha,\beta,\gamma\right]^T$ to dynamically fuse multiple retrieval signals.
To analyze the behavioral patterns of dynamic weights, we manually classified some queries and roughly divided them into the following three categories (not strictly mutually exclusive and are only used for analysis):
\begin{itemize}
    \item entity-centric query, targeting well-defined entities, such as a person, place, organization, or specific event;
    \item explanatory query, focusing on conceptual explanations, cause analysis, or high-level descriptions;
    \item multi-hop reasoning query, integrating multiple evidence segments or reasoning across documents to obtain the answer.
\end{itemize}

\begin{figure}
    \centering
    \includegraphics[width=\linewidth]{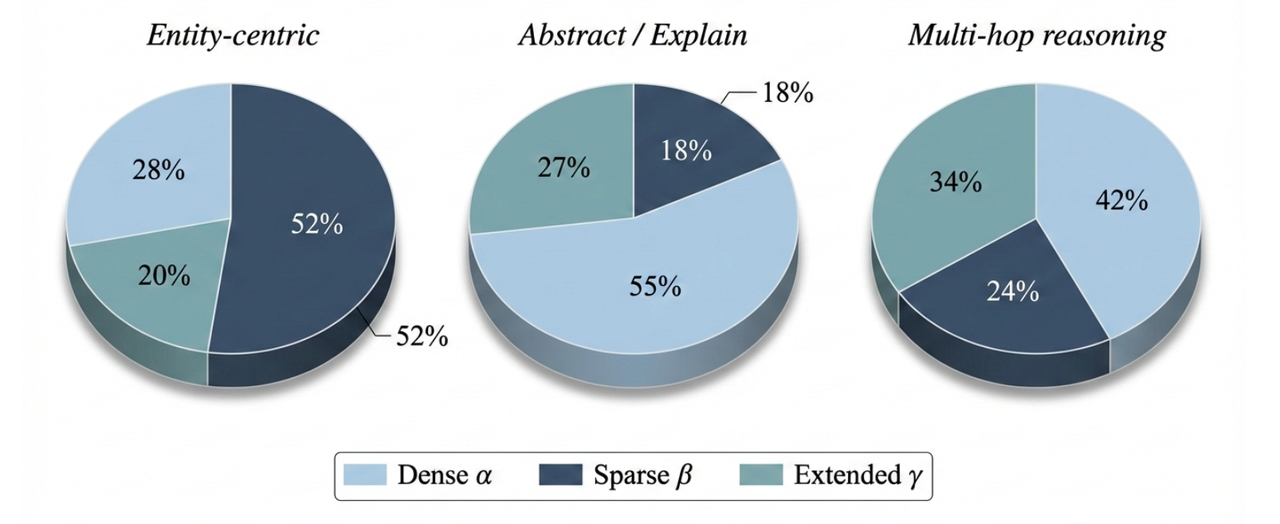}
    \caption{Distribution of channel weighting across different query types.}
    \label{fig:channel}
\end{figure}

\noindent
Figure~\ref{fig:channel} presents the weight distribution of retrieval channels for each query type.
We observe that entity-centric queries rely more on a sparse retrieval channel to improve term exact matching ability, and explanatory queries rely on dense retrieval to expand semantic coverage.
Meanwhile, multi-hop reasoning queries exhibit balanced weight among three retrieval channels to balance recall and semantic coherence.
These results indicate that dynamic channel weighting has an interpretable correspondence with the semantic features of the query.
Note that our weighting analysis does not provide strict guarantees on all queries due to the subjective classification of the query, where different annotation standards may lead to minor differences in weight distribution results.
Furthermore, the prediction results of dynamic weights rely on the SLM's understanding of query semantics, and its behavior may be influenced by model size and prompting methods. 
Nevertheless, we observed in the experiments that the overall trend remained consistent across different SLM settings.




\end{document}